\documentclass[sigconf]{acmart}
\usepackage{subcaption}
\usepackage{graphicx}
\usepackage{multirow}
\usepackage{amsmath}
\usepackage{xcolor}
\AtBeginDocument{%
  \providecommand\BibTeX{{%
    \normalfont B\kern-0.5em{\scshape i\kern-0.25em b}\kern-0.8em\TeX}}}

\copyrightyear{2022}
\acmYear{2022}
\setcopyright{rightsretained}
\acmConference[GECCO '22 Companion]{Genetic and Evolutionary
Computation Conference Companion}{July 9--13, 2022}{Boston, MA, USA}
\acmBooktitle{Genetic and Evolutionary Computation Conference
Companion (GECCO '22 Companion), July 9--13, 2022, Boston, MA, USA}
\acmDOI{10.1145/3520304.3529061}
\acmISBN{978-1-4503-9268-6/22/07}
\begin{document}

\title{Fair Feature Subset Selection using Multiobjective Genetic Algorithm}
\author{Ayaz Ur Rehman}
\email{ayaz.rehman@ndsu.edu}
\affiliation{%
  \institution{North Dakota State University}
  \city{Fargo}
  \state{North Dakota}
  \country{USA}
  \postcode{58102}
}

\author{Anas Nadeem}
\email{anas.nadeem@ndsu.edu}
\affiliation{%
 \institution{North Dakota State University}
 \city{Fargo}
 \state{North Dakota}
 \country{USA}
 \postcode{58102}
}

\author{Muhammad Zubair Malik}
\email{zubair.malik@ndsu.edu}
\affiliation{%
  \institution{North Dakota State University}
  \city{Fargo}
  \state{North Dakota}
  \country{USA}
  \postcode{58102}
}

\begin{abstract}
The feature subset selection problem aims at selecting the relevant subset of features to improve the performance of a Machine Learning (ML) algorithm on training data. Some features in data can be inherently noisy, costly to compute, improperly scaled, or correlated to other features, and they can adversely affect the accuracy, cost, and complexity of the induced algorithm. The goal of traditional feature selection approaches has been to remove such irrelevant features. In recent years ML is making a noticeable impact on the decision-making processes of our everyday lives. We want to ensure that these decisions do not reflect biased behavior towards certain groups or individuals based on the protected attributes such as age, sex, or race. In this paper, we present a feature subset selection approach that improves both fairness and accuracy objectives and computes Pareto-optimal solutions using the NSGA-II algorithm. We use statistical disparity as a fairness metric and F1-Score as a metric for model performance. Our experiments on the most commonly used fairness benchmark datasets with three different machine learning algorithms show that using the evolutionary algorithm we can effectively explore the trade-off between fairness and accuracy. 

\end{abstract}




\maketitle

\section{Introduction}
\label{sec:introduction}
Over the last decade, machine learning (ML) is becoming an integral part of many systems that make life-impacting decisions. In areas such as healthcare resource allocation, credit risk assessment, and recidivism predictions data-driven decisions are generated by machine learning algorithms. It is alarming that common users and decision-makers trust AI-based algorithms more than human experts, and underestimate the likelihood of errors inherent in such systems~\cite{AIAcceptance1}. These algorithms have the potential to amplify the biases that already exist in society and have serious legal and ethical implications. This has attracted researchers from diverse domains to explore various types of analysis that can provide insights into the development process of machine learning models. Our work focuses on this problem from the lens of feature selection and aims at providing a range of non-dominating choices to system engineers that give a trade-off between fairness and accuracy. This is important because fairness with respect to a specific attribute may or may-not be desired.

In this work, feature subset selection is addressed in the context of supervised machine learning, which is a subcategory of machine learning. Supervised learning uses labeled datasets to induce or train models to predict outcomes accurately. Dataset is composed of training instances, where each instance is described by a vector of feature (or attribute) values and a class label. As input data is fed into the model, the algorithm adjusts its weights until the model has been fitted according to some criteria which are usually accuracy. The key idea behind the feature selection approach is that the data contains some features that are either redundant or irrelevant and can thus be removed without incurring a loss of information and improving the compatibility of data with a learning model class ~\cite{JMLR:NEU}. We observe that publicly available datasets are biased and the data gathering process usually enhances inherent human biases~\cite{zhang2021ignorance}. Our key insight is that joint optimization for accuracy and fairness helps us remove features most likely associated with bias in data. In some sense, we want to lose information associated with human prejudice and bias encoded in feature space.

\section{Background}
\label{sec:background}



\subsection{Related Work in Fairness}
Fairness has gained much attention and the community has worked on different aspects of it. There has been some work done on detecting bias in a system that makes decisions for people. \textbf{Aequitas} \cite{saleiro2018aequitas} toolkit helps gauge fairness in a system using different metrics. \textbf{AI Fairness 360} \cite{bellamy2018ai} is another bias detection tool targeted towards solving the same problem. These tools play their part towards the end of the machine learning pipeline while in the post-processing phase resulting in helping the ML community and policymakers towards making better decisions~ \cite{mehrabi2021survey}.
Furthermore, researchers have developed different approaches to mitigate bias. An example is \textbf{Fairlearn} \cite{agarwal2018reductions} which uses reduction-based approaches to create a fair classifier. It not only detects, but also mitigates bias from the system. \textbf{AI Fairness 360} \cite{bellamy2018ai}  also offers similar services to remove bias. 

\subsection{Related Work in Feature Selection}
Genetic Algorithms have been used for the task of feature selection in the past \cite{babatunde2014genetic} and have performed much better than existing techniques which are confirmed from the work of \cite{oh2004hybrid, hamdani2006distributed}. NSGA-II \cite{deb2002fast} is one of the most efficient algorithms developed for multi-objective optimization. It has been used widely in feature selection as mentioned in \cite{singh2017optimal, soyel2011application}. \cite{babatunde2014genetic} used a similar technique using NSGA-II but with different objectives and they mention that multi-objective optimization is a really good approach for the task of feature selection. 


\begin{table*}
  \caption{Single Objective optimization}
  \label{tab: result1}
  \resizebox{\textwidth}{!}{%
  \begin{tabular}{|c|c|c|c|c|c|c|c|}
\hline
\textbf{Dataset} & \textbf{Model} & \textbf{Method} & \textbf{Accuracy} & \textbf{F1 score} & \textbf{Statistical Disparity Diff.} & \textbf{Total Features} & \textbf{Features selected} \\ \hline
German Credit Score & Logistic Regression & Naive Technique & 75\% & 0.833 & 0.0796 & 20 & - \\ \cline{3-8} 
 &  & Genetic Algorithm & 70\% & 0.82 & 0.00 & 20 & 6 \\ \cline{2-8} 
 & XGboost & Naive Technique & 74\% & 0.82 & 0.015 & 20 & - \\ \cline{3-8} 
 &  & Genetic Algorithm & 69\% & 0.80 & 0.00042 & 20 & 5 \\ \cline{2-8} 
 & SVM & Naive Technique & 70\% & 0.82 & 0.0 & 20 & - \\ \cline{3-8} 
 &  & Genetic Algorithm & 70\% & 0.82 & 0 & 20 & 13 \\ \hline
UCI Adult data set & Logistic Regression & Naive Technique & 80\% & 0.466 & 0.1077 & 14 & - \\ \cline{3-8} 
 &  & Genetic Algorithm & 76\% & 0.118 & 0.0092 & 14 & 7 \\ \cline{2-8} 
 & XGboost & Naive Technique & 87\% & 0.711 & 0.177 & 14 & - \\ \cline{3-8} 
 &  & Genetic Algorithm & 76\% & 0.0012 & 0.00069 & 14 & 2 \\ \cline{2-8} 
 & SVM & Naive Technique & 79\% & 0.26 & 0.02 & 14 & - \\ \cline{3-8} 
 &  & Genetic Algorithm & 76\% & 0.00042 & 0.0003 & 14 & 6 \\ \hline
\end{tabular}%
}
\end{table*}
\begin{table*}
  \caption{Bi-Objective optimization}
  \label{tab: result2}
  \resizebox{\textwidth}{!}{%
  \begin{tabular}{|c|c|c|c|c|c|c|}
\hline
\textbf{Dataset} & \textbf{Model} & \textbf{Objective} & \textbf{F1 score} & \textbf{Statistical Disparity Diff.} & \textbf{Total Features} & \textbf{Features Selected} \\ \hline
German Credit Score & Logistic Regression & Max Performance & 0.86 & 0.09 & 20 & 9 \\ \cline{3-7} 
 &  & Max Fairness & 0.82 & 0.00 & 20 & 4 \\ \cline{2-7} 
 & XGboost & Max Performance & 0.86 & 0.06 & 20 & 15 \\ \cline{3-7} 
 &  & Max Fairness & 0.81 & 732 x$ 10^{-6}$ & 20 & 13 \\ \cline{2-7} 
 & SVM & Max Performance & 0.86 & 0.045 & 20 & 9 \\ \cline{3-7} 
 &  & Max Fairness & 0.82 & 0.00 & 20 & 11 \\ \hline
UCI Adult data set & Logistic Regression & Max Performance & 0.466 & 0.1077 & 14 & 9 \\ \cline{3-7} 
 &  & Max Fairness & 0.118 & 0.0092 & 14 & 4 \\ \cline{2-7} 
 & XGboost & Max Performance & 0.719 & 0.177 & 14 & 10 \\ \cline{3-7} 
 &  & Max Fairness & 0.0012 & 692 x$ 10^{-6}$ & 14 & 3 \\ \cline{2-7} 
 & SVM & Max Performance & 0.56 & 0.18 & 14 & 4 \\ \cline{3-7} 
 &  & Max Fairness & 427 x$ 10^{-6}$ & 307 x$ 10^{-6}$ & 14 & 7 \\ \hline
\end{tabular}%
    }
\end{table*}

\begin{figure*}[t!]
    \centering
    \includegraphics[width=0.25\textwidth]{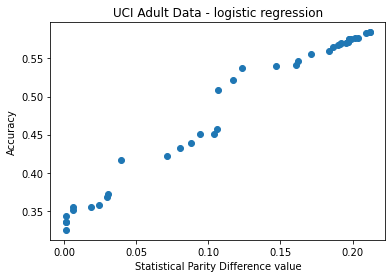}
    \hspace{1cm}
    \includegraphics[width=0.25\textwidth]{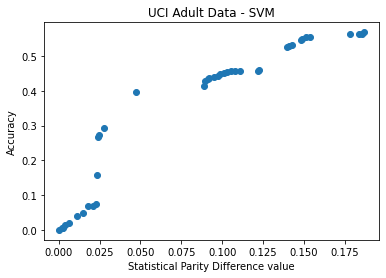}
    \hspace{1cm}
    \includegraphics[width=0.25\textwidth]{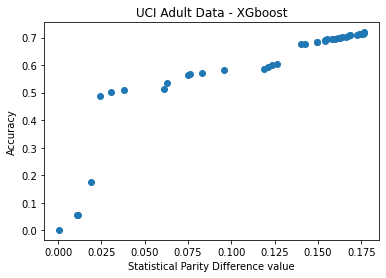}
    \hfill
    \includegraphics[width=0.25\textwidth]{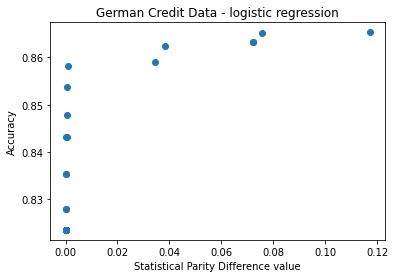}
    \hspace{1cm}
    \includegraphics[width=0.25\textwidth]{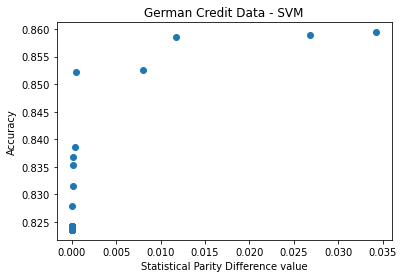}
    \hspace{1cm}
    \includegraphics[width=0.25\textwidth]{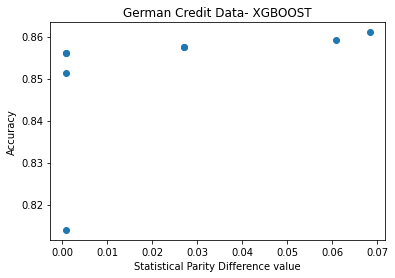}
    \caption{First front of the Pareto optimal solutions}
    \label{fig:img1}
\end{figure*}

\section{Methodology}
\label{sec:methodology}
\subsection{Overview}
The key idea is similar to existing wrapper-based approaches~\cite{kohavi1997wrappers}. The approach learns optimal features of a specific dataset with respect to fairness and F1-score on a specific ML algorithm. Features are encoded in chromosomes. The target algorithm is trained as part of fitness evaluation and estimated F1-score, and statistical disparity of induced algorithm on hold-out data is used to compute its fitness. NSGA-II algorithm is used to select the fittest chromes that encode optimal features. The two data-sets we have used are German data-set\footnote{\url{https://archive.ics.uci.edu/ml/datasets/statlog+(german+credit+data)}} and Census data-set\footnote{\url{https://archive.ics.uci.edu/ml/datasets/adult}}. As for the machine learning models, we have used Logisitic Regression~\cite{wright1995logistic}, Support Vector Machine~\cite{suthaharan2016support}, and XGboost~\cite{chen2016xgboost}. 

\subsection{Fariness Metric}
 We are using \textbf{Statistical Parity Difference} as the fairness metric. It is defined as the difference between the largest and the smallest group-level selection rate across all values of the sensitive feature(s). 
      \begin{equation}
            P(\hat{y} =m \mid G=m ) = P( \hat{y}=m \mid G=f ) 
      \end{equation}
    Here, \^ y denotes the decision generated by the model. G is a single protected attribute (race, sex, age, etc.). 

\subsection{Implementation Details}
\subsubsection{\textbf{Algorithm 1:}}\label{algo1}

Here we leverage the traditional genetic algorithm with few tailoring according to our problem requirement. 

\textit{Problem Encoding: } Our main idea is to represent our feature set as a string of binary values where "0" means that the respective feature is not part of the feature set, that is being fed to the machine learning model and vice versa. In experiment 1, where we are dealing with single-objective optimization, we start with a random population size of 40 chromosomes where each chromosome is a string of binary values. The length of the binary string is equal to the number of features in the original data set. For single-objective optimization, we used our own implementation of a genetic algorithm for feature selection. We give each feature an equal weight but along with this we also ensure that our protected feature, gender in our case, is part of our feature set throughout the process. 
We used the same configuration for all of our experiments. We fixed our generation size to be 40 chromosomes and we tested both of these data sets on three models. 

\textit{Genetic operators: }
 We used tournament selection to choose the best candidate from the population. We used the value of 0.5 as the cross-over rate and the value of 0.05 as the mutation rate.

\textit{Fitness function:}
 Our fitness function evaluates the statistical disparity difference value for each chromosome and uses it to generate a new population for the next generation.


\subsubsection{\textbf{Algorithm 2}} \label{algo2}
For our second algorithm, we leverage NSGA-II for the task of multi-objective optimization. Below we highlight the framework for the algorithm and present the experiment

\textit{Problem Encoding: }
We encoded our chromosome as in experiment 1 having binary values where value 0 or 1 represent respective feature is being included or not in our feature set, which is fed to the machine learning model.  We used Pymoo\footnote{https://pymoo.org/} minimize method which requires three arguments as input in order to minimize the different variables we give as input: Problem, Algorithm to be used for multi-objective optimization, and termination condition. The algorithm we used is NSGA-II. The configuration that we used for NSGA-II is shown below for reference: 

\begin{verbatim}
NSGA2(pop_size=40,sampling=get_sampling("bin_random"), 
      crossover=get_crossover("bin_hux"), 
      mutation=get_mutation("bin_bitflip"), 
      eliminate_duplicates=True)
\end{verbatim}

We defined our problem according to Pymoo standard which requires an evaluate method to be implemented. It simply requires that you define your fitness function inside that evaluate method and output an array of these scores for each individual chromosome.  NSGA-II uses Non dominated sort and crowding distance to rank the population in different set of classes. 
For complete detail of the NSGA-II refer to \cite{deb2002fast}.

\textit{Genetic operators: }
 We used tournament selection to choose the best candidate from the population. Similarly, we used the value of 0.5 as the cross-over rate and the value of 0.05 as the mutation rate.

\textit{Fitness functions: }
NSGA-II forms a set of fronts using Pareto dominance. The genetic operators(mutation and crossover) are performed using solutions from the first front. If the population size is not satisfied then the algorithm moves to the next front in order to complete the population. If the population size exceeds, while you are selecting candidates from a certain front, then the remaining members of that front are dropped. For the creation of these fronts, Pareto dominance and crowding distance are used to evaluate and categorize the members.


\subsection{Experimental Details}

\subsubsection{\textbf{Experiment 1}}

In the first experiment, our goal was to test the first algorithm, discussed in Section~\ref{algo1}, in order to find how traditional genetic algorithm performs in feature selection if there is only one objective to be optimized. In this experiment, we used fairness as the only objective to be optimized. As mentioned before that Fairness overall is a relative metric, so in our experiments, we are going with a single sensitive feature to be used in fairness metric calculation. Since this is a single objective optimization, it means we will get a single set of features which gives us ultimately the fairest results.

\subsubsection{\textbf{Experiment 2}}
In the second experiment, our goal was to observe how much evolutionary computational-based NSGA-II \cite{deb2002fast} algorithm, discussed in Section~\ref{algo2}, can perform over different data-sets using different models but in this case, we are providing two objectives to be optimized, Accuracy and Fairness. 
The two variables provided to NSGA-II to be optimized are F1 score and Statistical disparity. The result of our experiment represent the first front of the Pareto optimal solution for the data set after a machine learning model is applied.


\subsection{Experiment Results}
The results of our experiments are summarized and presented in Table \ref{tab: result1} and Table \ref{tab: result2}. Each table records evaluations done on three of the machine learning algorithms trained using two data sets, hence six experiments in total. Table \ref{tab: result1} lists accuracy, F1-score, and statistical disparity when fairness is the only objective to be optimized. Table \ref{tab: result2} lists maximum and minimum F1-score and statistical disparity when multi-objective optimization is performed using NSGA-II. As mentioned before, the results of fairness and accuracy are interpreted differently. The value for the statistical parity difference for the model closest to 0 is interpreted as least biased. Figure \ref{fig:img1}, shows the first front of the Pareto optimal solution for multi-objective optimization in our second experiment. Each graph in this figure represents a solution we obtained for two of the data sets and three machine learning models that we used in our experiments. Each point in the individual graph is a tuple that holds demographic disparity difference at its first index and F1-score at the second index. These points form a curve that shows us the trade-off pattern. It can be seen that as the statistical disparity difference decreases, the performance also decreases. One thing to be observed is that even though we used the same size population for each experiment, however, we did not get the same number of instances in the graphs. Some graphs have more points on optimal boundary than other.

The results of our first experiment are shown in Table \ref{tab: result1}. The results prove that the genetic algorithm is effective in this task of feature selection. The trade-off between accuracy and fairness is visible in this experiment. The genetic algorithm performs well on all three of the model and both data-sets. We observe a maximum decrease in statistical disparity difference in the case when we used the German credit score data set with logistics regression as the machine learning model. On the other hand, the minimum decrease is seen in the case when we use the German credit score data set and XGboost as the machine learning model. 

 Table \ref{tab: result2} highlights the results of our second experiment. The table gives us a holistic view of how the genetic algorithm actually performs in this experiment and we check both ends of the extremes. We got maximum fairness in the case with the UCI Adult data set with the SVM model and max performance with XGboost. An interesting observation from the results is that, for German credit score we got the same values for maximum performance for XGboost and SVM but different Statistical Disparity Differences. 
If we analyze the results from Table \ref{tab: result1} and Table \ref{tab: result2}, we can compare NSGA-II performance in terms of feature selection relative to traditional genetic algorithms. One clear observation is that we are getting almost the same maximum fairness for both of these techniques. The only case where NSGA-II performed much worse than our implementation is with the German credit score data set with the XGboost model. In all the other cases we see almost similar performance in terms of feature selection as well as Statistical Disparity Difference.
 \section{Discussion}

Machine Learning techniques are increasingly used to guide decisions in important societal spheres. However, there are growing concerns with regard to the validity of legal and ethical compliance of ML model’s valuations and predictions. In this work, we have explored feature selection using a fast sorting and elite multi-objective genetic algorithm, NSGA-II. While it is well known that there is a trade-off between fairness and accuracy objectives, however, to the best of our knowledge we are the first ones to present an approach that enables system engineers to explore it for a given training data and machine learning algorithm. There is an increased interest and a growing body of work in the area of fairness of ML and AI models, but a comprehensive understanding of this socio-technical issue is lacking. The work presented in this paper demonstrates that genetic and evolutionary approaches can significantly contribute to this domain.

\bibliographystyle{ACM-Reference-Format}
\bibliography{main}

\end{document}